\documentclass{article}

\usepackage{arxiv}

\usepackage[utf8]{inputenc} 
\usepackage[T1]{fontenc}    
\usepackage{hyperref}       
\usepackage{url}            
\usepackage{booktabs}       
\usepackage{amsfonts}       
\usepackage{nicefrac}       
\usepackage{microtype}      
\usepackage{lipsum}
\usepackage{graphicx}
\graphicspath{ {./images/} }

\usepackage{hyperref}
\usepackage{amsmath}
\usepackage{amsthm,amsmath,amsfonts,amssymb}
\usepackage{xcolor}
\usepackage{bm}
\usepackage{parskip} 
\usepackage{optidef}
\usepackage{multirow} 
\usepackage{graphicx}
\usepackage{standalone}
\usepackage{tikzscale}
\usepackage{subcaption} 

\usepackage{tikz}
\usetikzlibrary{positioning}

\usepackage{graphicx}

\ifLuaTeX
  \usepackage{selnolig}  
\fi
\usepackage[]{natbib}
\usepackage{bookmark}

\usepackage{mathtools}
\mathtoolsset{showonlyrefs} 

\definecolor{main}{HTML}{5989cf}    
\definecolor{sub}{HTML}{cde4ff}     

\usepackage[many]{tcolorbox}    	

\usepackage{setspace}               
\usepackage{multicol}               

\newtcolorbox{boxC}{
    colback = sub, 
    boxrule = 0pt  
}

\title{\bf {The Connection between Kriging and Large Neural Networks}}

\author{
 Marius Marinescu \\
  Engineering School of Fuenlabrada\\
  King Juan Carlos University \\
  Madrid, Spain \\
  \texttt{marius.marinescu@urjc.es} \\
}
\begin{document}
\maketitle
\begin{abstract}

AI has impacted many disciplines and is nowadays ubiquitous. In particular, spatial statistics is in a pivotal moment where it will increasingly intertwine with AI.
In this scenario, a relevant question is what relationship spatial statistics models have with machine learning (ML) models, if any.
%
In particular, in this paper, we explore the connections between Kriging and neural networks.
At first glance, they may appear unrelated. Kriging---and its ML counterpart, Gaussian process regression---are grounded in probability theory and stochastic processes, whereas many ML models are extensively considered \text{Black-Box} models. Nevertheless, they are strongly related. We study their connections and revisit the relevant literature.
The understanding of their relations and the combination of both perspectives may enhance ML techniques by making them more interpretable, reliable, and spatially aware.

\end{abstract}



\section{Introduction}\label{sec:1}



In recent decades, Artificial Intelligence (AI) and Machine Learning (ML) have revolutionized numerous scientific and engineering fields, permeating disciplines from computer vision to natural language processing. Spatial statistics, a cornerstone of geoscience, environmental modelling, and resource estimation, is no exception in this transformative wave. We are at a pivotal juncture where the integration of AI techniques with traditional spatial statistical methods promises to unlock new insights and capabilities. This convergence raises a fundamental question: what, if any, are the intrinsic relationships between classical spatial statistics models and modern ML architectures? In particular, how do time-honoured methods like Kriging—rooted in geostatistics and probabilistic frameworks—connect to the seemingly opaque ``black-box'' models of neural networks?

At first glance, Kriging and neural networks appear unrelated. Kriging, originally developed for mining applications to interpolate spatial data while accounting for autocorrelation and uncertainty, is deeply grounded in stochastic processes and linear prediction theory. Its machine learning counterpart, Gaussian Process Regression (GPR), extends this foundation into a full probabilistic paradigm. In contrast, neural networks, especially multilayer perceptrons (MLPs) and their deeper variants, are often viewed as flexible but non-interpretable function approximators driven by data and optimization algorithms. However, beneath these surface differences lies a profound mathematical kinship that bridges the two worlds.

This paper delves into these connections, emphasizing that the Kriging predictor is not only equivalent to the maximum a posteriori (MAP) GPR estimate but also emerges as a limiting case of large neural networks. To the best knowledge of the author Kriging was never presented from this machine learning perspective.

By exploring these links, we aim to foster a unified perspective that leverages the strengths of both domains: the interpretability and uncertainty quantification of spatial statistics with the scalability and universality of ML. Such an understanding may benefit ML models by infusing them with spatial awareness, while enriching spatial statistics with the adaptive power of neural architectures.

The remainder of the paper is organized as follows. First, Section ~\ref{sec:1} provides historical context on Kriging and GPR. Then, Section \ref{sec:2} elucidates the mathematical connection between Kriging and GPR. Next, Section \ref{sec:K_NN} establishes the convergence of MLPs to GPR in the infinite-width limit, with examples and numerical illustrations. Finally, we discuss broader implications, including extensions to deep networks or ML training, and conclude. 

\subsection{Kriging and GPR historical background}

In this section we provide a short historical introduction to Kriging and his ML counterpart, GPR.

\subsubsection{Kriging}


    

Daniel Gerhardus Krige worked in the gold mines of the Witwatersrand Basin, in South Africa \citep{krige2015}. One of the primary challenges in these mines was to accurately estimate the gold content of ore bodies. 

From a techno-economic perspective, a mine deserves to be exploited only if the cost of its extraction and processing does not exceed the value of the metal which can be extracted from it. The true grade, i.e. the amount of valuable material per unit of rock, of a panel\footnote{In the context of mining, a panel refers to a specific section or subdivision of a mine that is being worked on or has been prepared for extraction.} is not known before
its exploitation, so it is estimated by using a sample. At the beginning of the 1950s, the estimate was simply the average grade of the data
belonging to the panel or situated at its border. D. Krige noticed that gold deposits exhibited spatial continuity and that observations were not independent but correlated, and was struck by the fact that, on average, low-grade panels were underestimated and high-grade panels were overestimated. 




%

He observed that panel grade observations came close to the log-normal probability law (e.g. see \citet[Diagram 2]{krige1951}), and applied classical regression theory to improve and make unbiased the panel grade estimate, given the sample (e.g. see \citet[Diagram 24 and 25]{krige1962}). 
%
%
By using his method, the systematic error of using the sample average was tackled from an appropriate statistical perspective and the improvement of the accuracy in estimating the panel grades was high, see \citet[Sec. 6 - Improved Estimates Based on Statistical Theory]{krige1951}.

One of the reasons that motivated D. Krige work was the rudimentary use of statistics in mining at that time. I cite one of his statements from \citet{krige1951}:
``\textit{At present these methods consist almost 
entirely of the application of simple arithmetic and empirical formulae guided by practical experience and ignore the many advantage to be gained from a carefully statistical analysis ...}''
%
%
I refer to the previous papers for more details about D. Krige's concerns.

On the other hand, the (spatial) auto-correlation structure between observations was not taken into account, making estimators non-efficient. Here is where G. Matheron came into. 
%
The related French community started to use routinely the term `Le Krieage'  and G. Matheron coined the term `Kriging' in his publications in honour of D. Krige's work \citep{cressie1990}. 
He urged all scientists concerned with spatial interpolation to adopt this term and afterward became common in the Anglo-Saxon mining terminology.
What Kriging did and `Kriging' as coined by Matheron is not exactly the same. 
Given a random field, Matheron defined Kriging as a way to predict an unobserved value or a block average using the available observations. In particular, but without using this term, he derived the Best Linear Unbiased Predictor (BLUP) in the spatial statistics setting \citep{matheron1971}.
In the other hand, D. Krige original idea was to estimated a panel's true grade by a convex sum of the sample average of the whole data on the mine and the sample average of the panel. Thus, he used a weighted sum not of all observations (as Matheron), but of the global and local sample average. More details can be founds in~\cite{marinescu2024}.

Matheron is considered the founder of Geostatistics \citep{chiles2018}. In \citet{agterberg2004} he is presented as one of the greatest mathematical-statisticians from the twentieth century, at the standing of Ronald Fisher or John Tukey.

\subsubsection{GPR}

The formalisation and widespread use of GPR as we know it today, have different origins, and started in the 1990s, largely driven by advancements in the machine learning field. 

The concept of GPR is based on the concept of Gaussian Process (GP).  One of the earliest contributors to the theoretical foundations of GPs was Andrey Kolmogorov who laid the groundwork for the theory of stochastic processes \citep{kolmogorov1938}. In contrast to Kriging, 
whose development is more disparate, Kolmogorov’s work on probability theory and stochastic processes provided a rigorous mathematical framework that later proved fundamental to the development of GPR. A seminal  book about the mathematical foundation of GP, written by Kolmogorov's former students, is \citet{ibragimov1978}.


The use of GP in estimation emerged from the works of pioneering statisticians and mathematicians who sought to understand and model random processes over time and space.  In the latter half of the 20th century, researchers recognised the versatility of GPs in modelling different type of data. This period saw the development of key theoretical advancements and practical applications. 
The late 1990s and early 2000s marked a significant turning point for GPR, driven by the rise of machine learning and the increasing availability of computational resources.  

Christopher Williams and Carl Rasmussen were pivotal figures in the transition to machine learning. Their influential textbook called Gaussian Processes for Machine Learning  \citep{rasmussen2006}, synthesised previous theoretical developments and applied them to a wide range of problems, including regression and classification. They demonstrated the utility of GPR in providing not only solutions but also measures of uncertainty through the posterior distribution. Also, they discussed the connection of GPR to other mathematical and ML models \citep[Chapter 6 and 7]{rasmussen2006}. 



GPR has several appealing properties and advantages, which made it gain popularity:
\begin{itemize}

    \item A clear mathematical foundation, based on stochastic processes.
    
    \item The capacity to provide a measure of uncertainty on its predictions.
    \item The flexibility of the model. By choosing different kernels, GPR can model a wide range of target functions and capture different  data patterns.
    \item Its non parametric nature. Unlike parametric models, GPR does not assume a fixed form for the underlying (prior) functions, making it highly adaptable to complex datasets.
    \item GPR interpretability. The probabilistic nature and the explicit form of the covariance function make GPR interpretable, in opposition to black-box models.
\end{itemize}
The history of GPR is a testament of the interdisciplinary nature of modern statistical and machine learning methodologies. From its origins in the theoretical work of the early 20th century to its practical applications and its evolution into a fundamental tool in modern machine learning, GPR has continually adapted and expanded its scope. Today, it stands as a powerful and versatile method for modelling complex data.

\section{Kriging and GPR}\label{sec:2}

In this section, we describe the mathematical relation between Kriging and GPR. The GPR formalism will allow as to relate Kriging with Large Neural Networks (LNN).
Given a random field with observations, Kriging's objective is to predict an unobserved value or a block average. Kriging is based on the following equation:
\begin{equation*}
     Y(\bm{x}_i)=Z(\bm{x}_i)+\varepsilon_i, \ i=1, \ldots, n
\end{equation*}
where $Y(\bm{x}_i)$ are observations, $Z(\bm{x}_i)$ is a second order random field (not necessarily Gaussian) and $ \varepsilon_i $  are some i.i.d. noise. Let $m$ denote the mean function of $Z$ and $k$ the auto-covariance function.
To make a prediction, Kriging finds what is called in statistics the \text{BLUP}\footnote{The term BLUE is commonly used when estimating deterministic quantities, such as the parameters in classical regression. When estimating random quantities the term BLUP is used.}.
The well-known Simple-Kriging (SK) solution is (\cite{chiles2018}):
$$ \hat{Z}(\bm{x}_*)=m_* + \bm{k}_*^\top (\Sigma+ \sigma^2I)^{-1}(Y-\bm{m})$$

where $m_* = m(\bm{x_*})$, $\bm{m} = \big(m(\bm{x}_1),m(\bm{x}_2),...,m(\bm{x}_n)\big)^\top$,  $\bm{k}_*= \big(k(\bm{x}_1,\bm{x}_*), k(\bm{x}_2,\bm{x}_*),...,k(\bm{x}_n,\bm{x}_*)\big)^\top$, $\Sigma=\begin{pmatrix} k(\bm{x}_i,\bm{x}_j)
    \end{pmatrix}_{i,j=1,\dots,n}$ and $\sigma^2=\mathbb{V}[\varepsilon]$. 

As it can be observed, we are considering the SK description with a (possibly) non-zero mean but known. 
Now we present the GPR setting. The GPR formalism considers $Y(\bm{x})$ to be a \text{Gaussian} random field. Thus,
\begin{equation*}
    \big(Y(\bm{x}_1), \ldots, Y(\bm{x}_n)\big) \sim    \mathcal{N}(m(X),\ \Sigma+\sigma^2 I)
\end{equation*}
The estimation at a new location, $Z_* = Z(\bm{x}_*)$, is found by computing the conditional distribution $Z_* \mid Y=\bm{y}$:
\vspace{-1mm}

\begin{align*}
\Big[\begin{array}{c}
{Y} \\
Z_*
\end{array}\Big] & \sim \mathcal{N}\Big(\Big[\begin{array}{c} \bm{m} \\ m_*\end{array}\Big], \Big[\begin{array}{cc}
\Sigma +\sigma^2 I & \bm{k}_* \\
\bm{k}_*^\top & k_{**} 
\end{array}\Big]\Big)   & &\Rightarrow & 
Z_* \mid \bm{y} \sim \mathcal{N}\left(m_* + \bm{k}_*^\top (\Sigma+ \sigma^2I)^{-1} (\mathbf{y}-\bm{m}),\right. \\
&  &  &  &\left. k_{**}-\bm{k}_*^\top \Sigma^{-1} \bm{k}_*\right) .
\end{align*}

\vspace{-5mm}

where $k_{**}=k(\bm{x_*,\bm{x}_*)}$. The Maximum A Posteriori (\textbf{MAP}) of this conditional distribution yields the \text{SK} predictor and, moreover, it has the same variance (\cite{marinescu2024}).

Thus, \text{Kriging} prediction is the \text{same} as the \text{GPR MAP} predictor. In this sense, both techniques can be considered equivalent.


SK was used to describe the connection between Kriging and GPR. Other variants of Kriging such as Ordinary Kriging (OK) and Universal Kriging (UK) are also directly related with GPR.
The OK and UK methods use the so-called \text{Kriging system}, which is a clever set up that hides in the Lagrange multipliers the unknown mean function parameters, allowing in this way prediction. It can be shown that the procedure is \text{equivalent} to applying the Best Linear Unbiased Estimator (\text{BLUE}) method for the mean parameters and the \text{BLUP} method for the prediction.
For a complete discussion about the connections of SK, OK and UK with GPR, see \cite{marinescu2024}.


From now on, we use the terms Kriging and GPR interchangeably, as they are mathematically equivalent in the sense discussed earlier.

\section{Kriging as a multilayer perceptron}\label{sec:K_NN}

\begin{figure}[ht!]
    \centering
    \includegraphics[width=.6\linewidth]{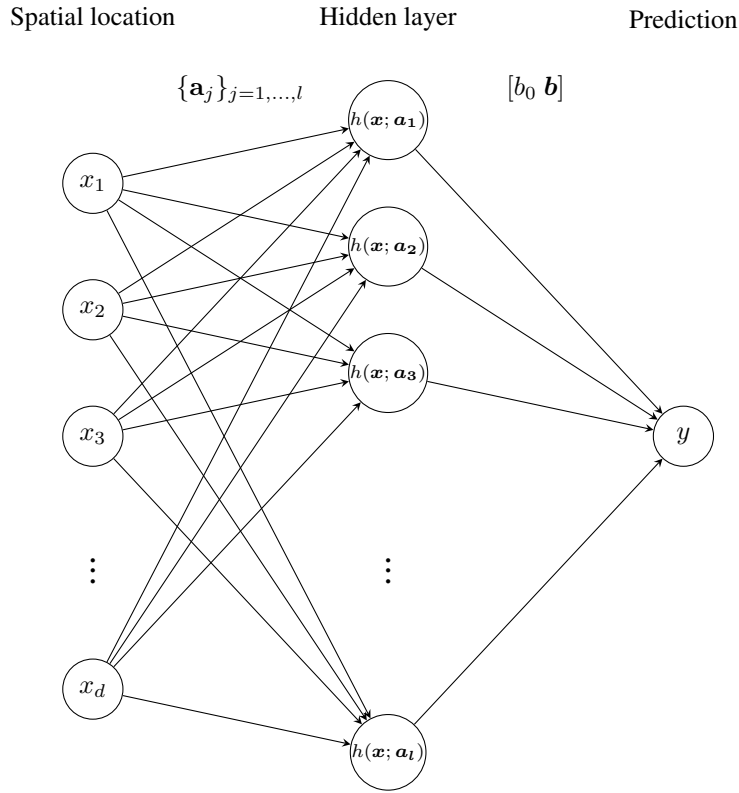}
    \caption{A MLP architecture with one hidden layer and a transfer function $h$. }
    \label{fig:NN}
\end{figure}

In this section, we show the connection that Kriging and a MLP with one hidden layer have. Consider a MLP which takes an input $\bm{x}$ (representing the spatial variables), has one hidden layer with $L$ neurons, and a transfer function $h$ (which we assume bounded) as illustrated on Fig. \ref{fig:NN}. Then, the output (representing the prediction) can be written as:
\begin{equation}\label{eq:NNoutput}
    y(\bm{x}) = b_0 + \sum_{j=1}^L b_j h(\bm{x}; \bm{a}_j)
\end{equation}
The output is a linear combination of the hidden units with a bias $b_0$.
This architecture has a very important property.
It has been shown that it can uniformly approximate any target continuous function on a compact set as the number of hidden neurons tends to infinity, provided the activation function is locally Riemann integrable and non-polynomial (\text{universal approximator}) (\cite{hornik1993}). An analogue property has also been reported for GPR models (\cite{micchelli2006universal, rasmussen2006}).

In order to make the Neural Network (NN) stochastic we introduce some randomness. Let the b's have independent zero-mean distribution of variance $\sigma^2_b$ and let each $\bm{a}_j$ be i.i.d. Then, $E[Y(\bm{x})] = 0$ by construction and the covariance is
\vspace{-3mm}
\begin{align}
        E[Y(\bm{x}) Y(\bm{x'})] &= \sigma^2_{{b}} + \sum_{j=1}^L \sigma^2_b  E[h(\bm{x}; \bm{a}_j)h(\bm{x'}; \bm{a}_j)] =  \notag \\
        &= \sigma^2_{{b}}  + L \cdot \sigma^2_b  E[h(\bm{x}; \bm{a)}h(\bm{x'}; \bm{a)}] \label{eq:cov_NN}
\end{align}

where in the first equation we have used the independence property, and the last equation follows because all of the hidden units are identically distributed. $Y$ is capitalised to empathise that now is random. 

It results that letting scale $L \cdot \sigma^2_b \to \text{cte.}$ on eq. \ref{eq:cov_NN} the covariance is bounded as $h$ has been assumed bounded.  
Interestingly, the stochastic process described in eq. \ref{eq:NNoutput} is an i.d.d. sum with bounded variance (eq. \ref{eq:cov_NN}) which, in virtue of the Central Limit Theorem (CLT), \text{converges} to a \text{Gaussian process} as the number of hidden units tends to infinity.
By evaluating $E[h(\bm{x}; \bm{a})h(\bm{x'}; \bm{a})]$, the explicit form of the covariance (and thus the \text{variogram}) that characterizes each NN can be obtained. Specifically,
 \begin{center}
%
\begin{equation}\label{eq:k_int}
    \overset{ \text{NN Cov}}{ \int_{\mathbb{R}^L} h(\bm{x}; \bm{a}) h(\bm{x'}; \bm{a}) f_{\bm{a}}(\bm{u}) \ d\bm{u} } \ \ \propto \overset{{ \text{Kriging Kernel}}}{  \textcolor{white}{\int} k(\bm{x}, \bm{x}') }
\end{equation}

 \end{center}

where $f$ is the density function of $\bm{a}$. 


Thus, Kriging and MLP models are strongly connected. Specifically, under the suppositions mentioned above, the MLP model tends to a Gaussian process when the number of hidden units tends to infinity. Furthermore, the mean function and covariance (kernel) of this GP, can directly be retrieved from the specification of the MLP model. 
Since the convergence is based on the CLT, typically 30 hidden units are usually enough to get an good approximation of the GP limit distribution. In Section~\ref{sec:numeric} numerical demonstrations are shown. For specific (theoretical) bounds the Berry-Esseen bounds may be used (\cite{vershynin2018}).


Due to works [5-6], among others, we introduce some examples of transfer functions and their corresponding covariance functions, which are listed in Table~\ref{tab:kernels}. It can be observed that well know kernels such as the linear, the squared exponential and the neural network have a corresponding MLP transfer function counterpart.
New stationary kernels may be produced as well by choosing different probability distributions for $\bm{a}$. 
For many transfer functions, it may be difficult to compute the integral  in eq. \ref{eq:k_int} explicitly in terms of elementary functions and numerical approximations may be used.
In general, most of the MLP transfer function produce non-stationary kernels (in Table~\ref{tab:kernels} only the Squared Exponential kernel is stationary). This may explain why NN architectures are so universal and can adapt well to many different applied problems.

Note that the so-called ``sigmoid kernel" in the literature, is never positive definite, and thus is not a valid covariance function (\cite{rasmussen2006}), leading to its omission.

    

\begin{table*}

\resizebox{1\linewidth}{!}{
\begin{tabular}{|c|c|c|c|}
    \hline
    Kernel Name  & Transfer Function &  Kernel & Kernel with $\bm{a}$ improper \\
    \hline
    \hline
    Linear &   $\begin{tabular}[c]{@{}l@{}}  $h(\bm{x}; \bm{a}) = \bm{a}^\top\bm{x}$ \\ {\small $\bm{a} \sim \mathcal{N} ( \bm{0}, \Sigma)$ } \end{tabular}$ &  $k(\bm{x}, \bm{x'}) = \bm{x}^\top \Sigma \bm{x}'$
    &  - \\ \hline
    Neural Network      & \begin{tabular}[c]{@{}l@{}} $h(\bm{x}; \bm{a}) = \text{erf}(\bm{a}^\top\bm{x})$ \\ {\small $\bm{a} \sim \mathcal{N} ( \bm{0}, \Sigma)$ } \end{tabular}& 
    \begin{tabular}[c]{@{}l@{}}$k(\bm{x}, \bm{x'}) = \frac{2}{\pi}\arcsin{\dfrac{2\bm{x}^\top \Sigma\bm{x}'}{(1 + 2\bm{x}^\top \Sigma\bm{x})(1 + 2\bm{x}'^\top \Sigma\bm{x}') }} $ \\ \end{tabular} 
    &  $k(\bm{x}, \bm{x'}) = \frac{2}{\pi}\arcsin{\dfrac{\bm{x}^\top \bm{x}'}{||\bm{x}||^2 ||\bm{x}'||^2 }} $ \\ \hline
    Squared Exponential &
    \begin{tabular}[c]{@{}l@{}}  $h(\bm{x}; \bm{a}) = \sqrt{2}\cos{(\bm{a}^\top \bm{x}+\varphi_0)} $ \\ {\small $\bm{a} \sim \mathcal{N} ( \bm{0}, \frac{1}{\sigma^2}I), \ \ \varphi_0 \sim \mathcal{U}{(0,2\pi)} $} \end{tabular} &
    $k(\bm{x}, \bm{x'}) = \exp \left(-\frac{\left||\mathbf{x}-\mathbf{x}^{\prime}\right||^2}{2\sigma^2}\right)$   & $k(\bm{x}, \bm{x'}) = \begin{cases}
         1 & \text{ if } \bm{x}=\bm{x}' \\
         0 & \text{ if } \bm{x}\neq\bm{x}'
    \end{cases}$ \\
    \hline
     \begin{tabular}[c]{@{}l@{}} Non-stationary \\ Squared Exponential  \end{tabular}  &  
     \begin{tabular}[c]{@{}l@{}}  $h(\bm{x}; \bm{a}) = \exp{(-||\bm{x} - \bm{a}||^2 / 2\sigma_g^2)}$ \\ {\small $\bm{a} \sim \mathcal{N} ( \bm{0}, \sigma_a^2I)$}  \end{tabular} & 
     \begin{tabular}[c]{@{}l@{}} $k(\bm{x}, \bm{x'})  \propto \exp \left(-\frac{\mathbf{x}^{\top} \mathbf{x}}{c_1}\right) \exp \left(-\frac{\left||\mathbf{x}-\mathbf{x}^{\prime}\right||^2}{c_2}\right) \exp \left(-\frac{\mathbf{x}^{\prime \top} \mathbf{x}^{\prime}}{c_1}\right)$ \\ {\small for some cte. $c_1$, $c_2$}. \end{tabular}  &   
     $k(\bm{x}, \bm{x'}) \propto \exp \left(-\frac{\left||\mathbf{x}-\mathbf{x}^{\prime}\right||^2}{4\sigma_g^2}\right)$  \\ \hline
     Arc-cosine I  & 
     \begin{tabular}[c]{@{}l@{}}  $h(\bm{x}; \bm{a}) = \delta_{\text{Heaviside}}(\bm{a}^\top\bm{x}) $ \\ {\small $\bm{a} \sim \mathcal{N} ( \bm{0}, I)$ } \end{tabular} &
     $k(\bm{x}, \bm{x'}) = 1 - \frac{\theta}{\pi}, \ \ \cos{\theta} = \frac{\bm{x}^\top \bm{x'}}{||\bm{x}|| \cdot ||\bm{x'}||}$ & - \\ \hline
     Arc-cosine II  & 
     \begin{tabular}[c]{@{}l@{}}  $h(\bm{x}; \bm{a}) = \max(0, \bm{a}^\top\bm{x}) \ - \text{ ReLU} $ \\ {\small $\bm{a} \sim \mathcal{N} ( \bm{0}, I)$ } \end{tabular} &
    $k(\bm{x}, \bm{x'}) = \frac{||\bm{x}|| \cdot||\bm{x'}||}{\pi}(\sin{\theta}+(\pi-\theta)\cos{\theta}), \ \ \cos{\theta} = \frac{\bm{x}^\top \bm{x'}}{||\bm{x}|| \cdot ||\bm{x'}||}$ & - \\ 
    \hline
\end{tabular}
}
\caption{List of kernels and transfer functions (non-extensive). The Neural Network and Non-stationary Squared Exponential kernels can be found in \cite{williams1998}. The Arc-Cosine I and II can be found in \cite{cho2009}. The calculation of the Linear and Squared Exponential kernels, which were not found in the literature, are given in the Appendices \ref{ap:1} and \ref{ap:2}.} \label{tab:kernels}
   
\end{table*}

\subsection{Some numerical demonstrations}\label{sec:numeric}

In this section, we test numerically the theoretic relation described in Section~\ref{sec:K_NN} for particular cases. Specifically, we simulate from a MLP and a Gaussian Process and perform a statistical test on the stochastic samples to assess whether they can be considered equally distributed or not.

\begin{figure*}[ht]
    \centering
    \begin{subfigure}[t]{0.48\textwidth}
        \centering
        \includegraphics[width=\linewidth, trim=3.5cm 9.5cm 3.5cm 10cm, clip]{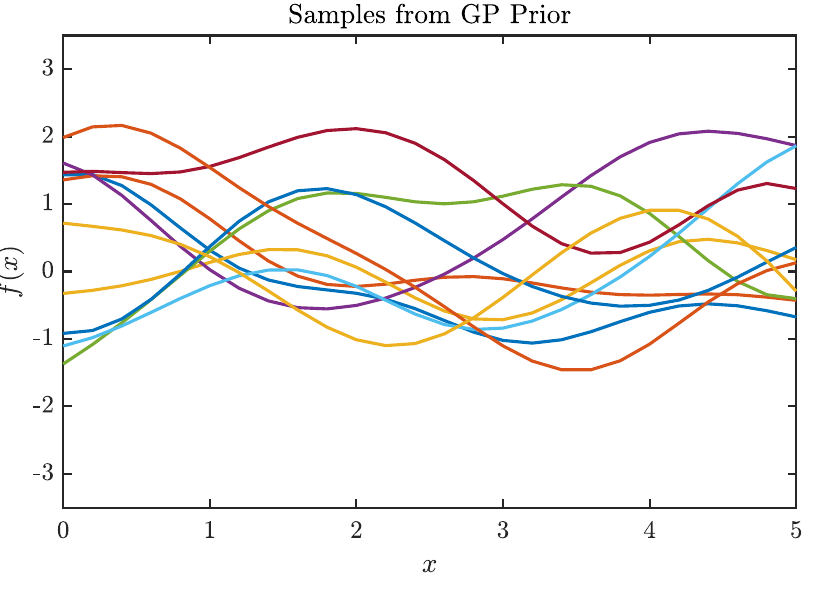}
        \caption{GP prior with squared exponential kernel.}
        \label{fig:gp_prior_rbf}
    \end{subfigure}
    \hfill
    \begin{subfigure}[t]{0.48\textwidth}
        \centering
        \includegraphics[width=\linewidth, trim=3.5cm 9.5cm 3.5cm 10cm, clip]{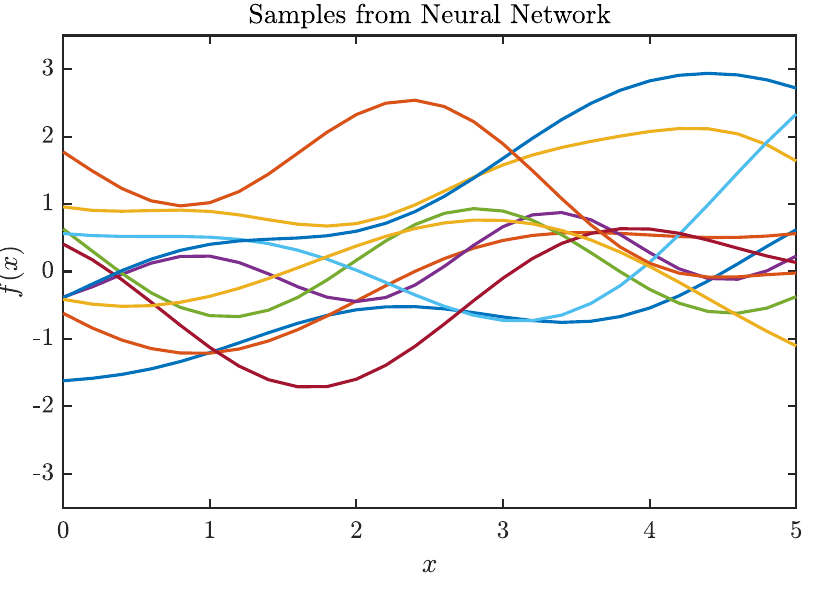}
        \caption{MLP with cosine activation function and 20 hidden units.}
        \label{fig:gp_prior_matern}
    \end{subfigure}
    \caption{Samples from GP and NN. The length scale $\sigma$ is chosen to be 1.}
    \label{fig:gpnn_priors}
\end{figure*}

In Figure~\ref{fig:gpnn_priors}, a pair of 10 sample paths are shown. The figure on the left is generated from a GP using the squared exponential kernel, while the figure on the right is generate using the MLP model with the cosine activation function. 

%
To test whether they came from the same distribution we have used a rank test for random function, based on the concept of depth of functional data \cite[Section 8]{lopez2009}. The test uses the concept of band delimited by the graph of two or more functions from the sample. Using this tool a measure of depth or centrality of each random function is established, providing a natural centre-outward ordering of them. This has several application, including a rank test\footnote{This rank test can be viewed as a generalization applied to random functions of the two-sided Wilcoxon signed-rank test.} to decide whether two groups of curves comes from the same population (see \cite{lopez2009} for more information). The null hypothesis is that they comes from the same population whereas the alternatives hypothesis is they don't.

To apply the test, we have generated 50 random realisations from the linear and squared exponential stochastic processes by means of both a MLP model and a GP model. For the MLP model, only 200 hidden units were used, and in both cases the covariance matrix of $\bm{a}$ was chosen to be the identity matrix. Applying the test, it results that the null hypothesis shall be accepted in both cases, with $p$-values of 0.63 and 0.91, respectively. 
%
Other configurations varying the sample size, number of hidden units, and variances were also tested, yielding similar results. 
%
Hence, these tests empirically support the theoretical results discussed in this section.

\subsection{Further connections}

There are \text{generalizations} of the connection between GPs and MLP to Deep Neural Networks (DNNs) and Convolutional Neural Networks (CNNs). 
In the infinite-width limit, in which the number of neurons per hidden layer tends to infinity with a suitably scaled random initialization, DNN and CNN converges to a GP. This limit GP has a kernel determined by the network architecture, known as the Neural Network Gaussian Process (NNGP) kernel (\cite{lee2017deep,garriga2018deep}). Moreover, the equivalence of NNs and GPs in the infinite-width limit, has been generalised and described for a wide class of NN architectures in \cite{yang2019}.

The common training of NN models has also been related to the GPR estimation. The so-called Neural Tangent Kernel (NTK) is a kernel that describes the evolution of deep artificial neural networks during their training by gradient descent (\cite{jacot2018, lenaic2019lazy}).
In the infinite-width regime, gradient descent on network parameters induces a linearized training dynamics governed by the NTK, which itself defines a kernel analogous to the covariance in a GP. Interestingly, this property reveals that DNN training can be interpreted as performing kernel regression in a high-dimensional feature space, bridging the conceptual gap between kernel methods training and deep learning.

GPR is also related with Support Vector Machines (SVMs) and Relevance Vector Machines (RVMs) \cite[Chapter 6]{rasmussen2006}, yet they are not equivalent. The former models typically yields more sparse solution, and differs in their predictive distribution.




In conclusion, Kriging and GPR sits at the intersection of several key machine learning paradigms, revealing deep conceptual links. 
This make clear that they are not an isolated methodology but rather part of a broader landscape linking probabilistic inference, kernel methods, and deep learning. 
%
Ultimately, the kernel-based perspective can enhance ML techniques making them more interpretable, reliable, and spatially aware.





\bibliography{bibliography.bib}

\section*{Appendix}

\begin{appendix}
    \section{Linear kernel}\label{ap:1}

    We will show that $$E[h(\bm{x}; \bm{a)}h(\bm{x}'; \bm{a)}] = \bm{x}^\top \Sigma \bm{x}',$$ 
    when $\bm{a} \sim \mathcal N(0,\Sigma)$ and $h$ linear.

    \vspace{2mm}
    
    Demonstration
    \begin{align*}
         \hspace{-.9cm}    E[h(\bm{x}; \bm{a)}h(\bm{x}'; \bm{a)}] &= E[\bm{a}^\top \bm{x} \cdot \bm{a}^\top \bm{x}'] = E[\bm{x}^\top \bm{a} \cdot \bm{a}^\top \bm{x}'] = \\
            &=\bm{x}^\top E[  \bm{a} \cdot \bm{a}^\top ] \bm{x}' = \boxed{\bm{x}^\top \Sigma \bm{x}'}
    \end{align*}
    where we have used that $\bm{a}^\top \bm{x} = \bm{x}^\top \bm{a}$ since it is a scalar and that $E[  \bm{a} \cdot \bm{a}^\top ]$ is the covariance of $\bm{a} $ by definition.

\section{Stationary squared exponential kernel}\label{ap:2}

    We will show that $$E[h(\bm{x}; \bm{a)}h(\bm{x}'; \bm{a)}] = \exp\!\left(-\frac{\|\bm{x}-\bm{x}'\|^2}{2\sigma^2}\right),$$ 
    when $\bm{a} \sim \mathcal N(0,\Sigma=\sigma^{-2}I)$, $ \varphi_0 \sim \mathcal{U}{(0,2\pi)}$ and $h(\bm{x}; \bm{a)} = \sqrt{2}\cos{(\bm{a}^\top \bm{x}+\varphi_0)}$.

    \vspace{2mm}
    
    Demonstration
    \begin{align*}
        \hspace{-.9cm} E[h(\bm{x}; \bm{a)}h(\bm{x}'; \bm{a)}] = E[\sqrt{2}\cos{(\bm{a}^\top \bm{x}+\varphi_0)} \sqrt{2}\cos{(\bm{a}^\top \bm{x}'+\varphi_0)}]
    \end{align*}
    Now we use the trigonometric identity: 
    $$ \cos{a}\cos{b}=\frac{1}{2} (\cos{(a-b)} + \cos{(a+b)})$$
    resulting in:
    \begin{align*}
       & \hspace{-.9cm}  E[h(\bm{x}; \bm{a)}h(\bm{x}'; \bm{a)}] = \\
        &E[\cos{(\bm{a}^\top (\bm{x}- \bm{x}'))} ] + E[\cos{(\bm{a}^\top (\bm{x}+ \bm{x}')) + 2\varphi_0} ] = \\
        &E[\cos{(\bm{a}^\top (\bm{x}- \bm{x}'))} ]
    \end{align*}
    The second expectation vanishes since it is an average of a random phase: $\int_0^{2\pi} \cos{(2\varphi_0 + cte.)}  \cdot \mathcal{U}{(0,2\pi)} \ d\varphi_0 = 0$. 

    The remaining expectation is only dependant of $\bm{a}$. Now we use that $ {E}\!\left[\cos\!\big(\bm{a}^\top \bm{t}\big)\right]
    = \text{Re} \big(\,{E}\big[e^{i\bm{a}^\top \bm{t}}\big] \big)$ and that the characteristic function of  $\bm{a} \sim \mathcal N(0,\Sigma)$ is 
    $${E}\!\left[e^{i \bm{a}^\top \bm{t}}\right] \;=\; \exp\!\left(-\tfrac12 \bm{t}^\top \Sigma \bm{t}\right).$$

Setting \(\bm{t}=\bm{x}- \bm{x}'\) and taking the real part (which is already real and positive) gives
\[
{E}_{\bm{a},\varphi_0}\big[h(\bm{x}; \bm{a)}h(\bm{x}'; \bm{a)}\big]
= \exp\!\left(-\tfrac12 (\bm{x}-\bm{x}')^\top \Sigma (\bm{x}-\bm{x}')\right).
\]

In particular, for \(\Sigma=\sigma^{-2}I\)
\[
\boxed{\;
\mathbb{E}_{\bm{a},\varphi_0}\big[h(\bm{x}; \bm{a)}h(\bm{x}'; \bm{a)}\big]
= \exp\!\left(-\frac{\|\bm{x}-\bm{x}'\|^2}{2\sigma^2}\right)
\;}_{\ .}
\]

\end{appendix}
\end{document}